\documentclass{bmvc2k}

\usepackage{hyperref}       % hyperlinks
\usepackage{url}            % simple URL typesetting
\usepackage{nicefrac}       % compact symbols for 1/2, etc.
\usepackage{amsfonts}       % blackboard math symbols
\usepackage{amsmath,amssymb}
\usepackage{algorithm, algpseudocode}
\usepackage{multirow}
\usepackage{caption}
\usepackage{array}
\title{Learning Accurate Low-Bit Deep Neural Networks with Stochastic Quantization}

\addauthor{Yinpeng Dong}{dyp17@mails.tsinghua.edu.cn}{1}
\addauthor{Renkun Ni}{rn9zm@virginia.edu}{2}
\addauthor{Jianguo Li}{jianguo.li@intel.com}{3}
\addauthor{Yurong Chen}{yurong.chen@intel.com}{3}
\addauthor{Jun Zhu}{dcszj@mail.tsinghua.edu.cn}{1}
\addauthor{Hang Su}{suhangss@mail.tsinghua.edu.cn}{1}

\addinstitution{
 Department of Computer Science and Technology, 
 Tsinghua University\\
 Beijing, China
}
\addinstitution{
 University of Virginia
}
\addinstitution{
 Intel Labs China\\
 Beijing, China
}

\runninghead{Dong \etal}{Stochastic Quantization}

\def\eg{\emph{e.g}\bmvaOneDot}

\def\etal{\emph{et al}\bmvaOneDot}

\begin{document}

\maketitle

\begin{abstract}
Low-bit deep neural networks (DNNs) become critical for embedded applications due to their low storage requirement and computing efficiency.
However, they suffer much from the non-negligible accuracy drop.
This paper proposes the stochastic quantization (SQ) algorithm for learning accurate low-bit DNNs.
The motivation is due to the following observation.
Existing training algorithms approximate the real-valued elements/filters with low-bit representation all together in each iteration.
The quantization errors may be small for some elements/filters, while are remarkable for others,
which lead to inappropriate gradient direction during training, and thus bring notable accuracy drop.
Instead, SQ quantizes a portion of elements/filters to low-bit with a stochastic probability inversely proportional to the quantization error, while keeping the other portion unchanged with full-precision.
The quantized and full-precision portions are updated with corresponding gradients separately in each iteration.
The SQ ratio is gradually increased until the whole network is quantized.
This procedure can greatly compensate the quantization error and thus yield better accuracy for low-bit DNNs.
Experiments show that SQ can consistently and significantly improve the accuracy for different low-bit DNNs on various datasets and
various network structures.
%In particular, we show that the proposed algorithm can even beat full-precision DNNs in accuracy with only 2-bits DNNs.
%This is by far the first algorithm could achieve to the best of our knowledge.
\end{abstract}

%-------------------------------------------------------------------------
\section{Introduction}
Deep Neural Networks (DNNs) have demonstrated significant performance improvements in a wide range of computer vision tasks~\cite{DeepReview_Lecun_2015}, such as image classification~\cite{krizhevsky2012imagenet,simonyan2014very,szegedy2015going,he2015deep}, object detection~\cite{girshick2014rich,ren2015faster} and semantic segmentation~\cite{long2015fully,chen2014semantic}.
DNNs often stack tens or even hundreds of layers with millions of parameters to achieve the promising performance.
Therefore, DNN based systems usually need considerable storage and computation power.
This hinders the deployment of DNNs to some resource limited scenarios, especially low-power embedded devices in the emerging Internet-of-Things (IoT) domain.

Many works have been proposed to reduce model parameter size or even computation complexity due to the high redundancy in DNNs~\cite{denil2013predicting,han-learning}.
Among them, low-bit deep neural networks~\cite{courbariaux2015binaryconnect,rastegari2016xnor,li2016ternary, zhou2016dorefa,zhu2016trained,binarynet,hubara2016quantized}, which aim for training DNNs with low bitwidth weights or even activations, attract much more attention due to their promised model size and computing efficiency.
In particular, in BinaryConnect (BNN)~\cite{courbariaux2015binaryconnect}, the weights are binarized to $+1$ and $-1$ and multiplications are replaced by additions and subtractions to speed up the computation.
In~\cite{rastegari2016xnor}, the authors proposed binary weighted networks (BWN) with weight values to be binarized plus one scaling factor for each filter,
and extended it to XNOR-Net with both weights and activations binarized.
Moreover, Ternary Weight Networks (TWN)~\cite{li2016ternary} incorporate an extra $0$ state, which converts weights into ternary values \{$+1$, $0$, $-1$\} with $2$-bits width.
However, low-bit DNNs are challenged by the non-negligible accuracy drop, especially for large scale models (\eg, ResNet~\cite{he2015deep}).
We argue that the reason is due to that they quantize the weights of DNNs to low-bits all together at each training iteration.
The quantization error is not consistently small for all elements/filters.
It may be very large for some elements/filters, which may lead to inappropriate gradient direction during training, and thus makes the model converge to relatively worse local minimum.

Besides training based solutions for low-bit DNNs, there are also post-quantization techniques, which focus on directly quantizing the pre-trained full-precision models \cite{lin2015fixed, miyashita2016logquant, zhou2017incremental}. These methods have at least two limitations.
First, they work majorly for DNNs in classification tasks, and lack the flexibility to be employed for other tasks like detection, segmentation, etc.
Second, the low-bit quantization can be regarded as a constraint or regularizer~\cite{courbariaux2015binaryconnect} in training based solutions towards a local minimum in the low bitwidth weight space. However, it is relatively difficult for post-quantization techniques (even with fine-tuning) to transfer the local minimum from a full-precision weight space to a low bitwidth weight space losslessly.

In this paper, we try to overcome the aforementioned issues by proposing the Stochastic Quantization (SQ) algorithm to learn accurate low-bit DNNs.
Inspired by stochastic depth \cite{stochastic} and dropout \cite{dropout}, the SQ algorithm stochastically select a portion of weights in DNNs and quantize them to low-bits in each iteration, while keeping the other portion unchanged with full-precision. The selection probability is inversely proportional to the quantization error.
%The purpose of this stochastic selection and quantization is to avoid using inaccurate gradient information.
We gradually increase the SQ ratio until the whole network is quantized.

We make a comprehensive study on different aspects of the SQ algorithm.
\textit{First}, we study the impact of selection granularity, and show that treating each filter-channel as a whole for stochastic selection and quantization performs better than treating each weight element separately.
The reason is that the weight elements in one filter-channel are interacted with each other to represent a filter structure,
so that treating each element separately may introduce large distortion when some elements in one filter-channel are low-bits while the others remain full-precision.
\textit{Second}, we compare the proposed \textit{roulette} algorithm to some deterministic selection algorithms.
The roulette algorithm selects elements/filters to quantize with the probability inversely proportional to the quantization error.
If the quantization error is remarkable, we probably do not quantize the corresponding elements/filters, because they may introduce inaccurate gradient information.
The \textit{roulette} algorithm is shown better than deterministic selection algorithms since it eliminates the requirement of finding the best initial partition, and has the ability to explore the searching space for a better solution due to the exploitation-exploration nature of stochastic algorithms.
\textit{Third}, we compare different functions to calculate the quantization probability on top of quantization errors.
\textit{Fourth}, we design an appropriate scheme for updating the SQ ratio.

%Beyond quantization, stochastic partition also acts as a regularizer during training to avoid overfitting, where we stochastically quantize a group of weights rather than mask a group of weights in Dropout~\cite{srivastava2014dropout}.

The proposed SQ algorithm is generally applicable for any low-bit DNNs including BWN, TWN, etc.
Experiments show that SQ can consistently improve the accuracy for different low-bit DNNs (binary or ternary) on
various network structures (VGGNet, ResNet, etc) and various datasets (CIFAR, ImageNet, etc).
For instance, TWN trained with SQ can even beat the full-precision models in several testing cases.
Our main contributions are:
\begin{itemize}
\addtolength{\itemsep}{-0.05in}
\item[(1)] We propose the stochastic quantization (SQ) algorithm to overcome the accuracy drop issue in existing low-bit DNNs training algorithms.
\item[(2)] We comprehensively study different aspects of the SQ algorithm such as selection granularity, partition algorithm, definition of quantization probability functions, and scheme for updating the SQ ratio, which may provide valuable insights for researchers to design other stochastic algorithms in deep learning.
\item[(3)] We present strong performance improvement with the proposed algorithm on various low bitwidth settings, network architectures, and benchmark datasets.
\end{itemize}

\section{Low-bit DNNs}
In this section, we briefly introduce some typical low-bit DNNs as prerequisites, including BNN~\cite{courbariaux2015binaryconnect}, BWN~\cite{rastegari2016xnor} and TWN~\cite{li2016ternary}.
Formally, we denote the weights of the $l$-th layer in a DNN by $\mathcal{W}_l = \{\mathbf{W}_1, \cdots,\mathbf{W}_i,\cdots,\mathbf{W}_{m}\}$, where $m$ is the number of output channels, and $\mathbf{W}_i \in \mathbb{R}^d$ is the weight vector of the \textit{i}-th filter channel,
in which $d = n\times w \times h$ in conv-layers and $d = n$ in FC-layers ($n$, $w$ and $h$ represent \textit{input channels}, \textit{kernel width}, and \textit{kernel height} respectively). $\mathcal{W}_l$ can also be viewed as a weight matrix with $m$ rows, and each row $\mathbf{W}_i$ is a $d$-dimensional vector.
For simplicity, we omit the subscript $l$ in the following.

\vspace{1ex}
\noindent\textbf{BinaryConnect} uses a simple stochastic method to convert each 32-bits weight vector $\mathbf{W}_i$ into binary values $\mathbf{B}_i$ with the following equation
\begin{equation}
\footnotesize
\mathbf{B}_i^j =
  \begin{cases}
    +1  &     \quad \text{with probability } p = {\sigma}(\mathbf{W}_i^j), \\
    -1  &     \quad \text{with probability } 1 - p.
  \end{cases}
\end{equation}
where ${\sigma}(x) = \max(0, \min(1, \frac{x + 1}{2}))$ is the \textit{hard sigmoid} function, and \textit{j} is the index of elements in $\mathbf{W}_i$.
During the training phase, it keeps both full-precision weights and binarized weights. Binarized weights are used for gradients and forward loss computation, while full-precision weights are applied to accumulate gradient updates.
During the testing phase, it only needs the binarized weights so that it reduces the model size by $32\times$.

\vspace{1ex}
\noindent\textbf{BWN} is an extension of BinaryConnect, but introduces a real-valued scaling factor $\alpha \in \mathbb{R}^+$ along with $\mathbf{B}_i$ to approximate the full-precision weight vector $\mathbf{W}_i$ by solving an optimization problem $\mathcal{J} = \min \|\mathbf{W}_i - \alpha \mathbf{B}_i\|$ and obtaining:
\begin{equation}
\footnotesize
    \mathbf{B}_i = \mathrm{sign}(\mathbf{W}_i) \quad \text{and} \quad \alpha = \frac{1}{d}\displaystyle\sum\nolimits_{j=1}^{d}|\mathbf{W}_i^j|.
\end{equation}

\vspace{1ex}
\noindent\textbf{TWN} introduces an extra 0 state over BWN and approximates the real-valued weight vector $\mathbf{W}_i$ more accurately with a ternary value vector $\mathbf{T}_i \in \{1, 0, -1\}^d$ along with a scaling factor $\alpha$, while still keeping high model-size compression (16$\times$).
It solves the optimization problem $\mathcal{J}=\min \|\mathbf{W}_i - \alpha \mathbf{T}_i\|$ with an approximate solution:
\begin{equation}
\footnotesize
\mathbf{T}_i^j =
   \begin{cases}
     +1 &  \text{if }\,\, \mathbf{W}_i^j > \Delta \\
     0 &  \text{if }\,\, |\mathbf{W}_i^j| \leq \Delta \\
     -1 &  \text{if }\,\, \mathbf{W}_i^j < -\Delta \\
   \end{cases}
\quad \text{and} \quad \alpha = \frac{1}{|\mathbf{I}_{\Delta}|}\displaystyle\sum_{i\in \mathbf{I}_{\Delta} }|\mathbf{W}_i^j|,
\end{equation}
where $\Delta$ is a positive threshold with following values
\begin{equation}
\footnotesize
    \Delta = \frac{0.7}{d}\displaystyle\sum\nolimits_{j=1}^{d}|\mathbf{W}_i^j|,
\end{equation}
$\mathbf{I}_{\Delta} = \{j\mid|\mathbf{W}_i^j| > \Delta\}$ and $|\mathbf{I}_\Delta|$ denotes the cardinality of set $\mathbf{I}_\Delta$.

\section{Method}
\begin{figure}[]
\centering
  \includegraphics[width=0.99\textwidth]{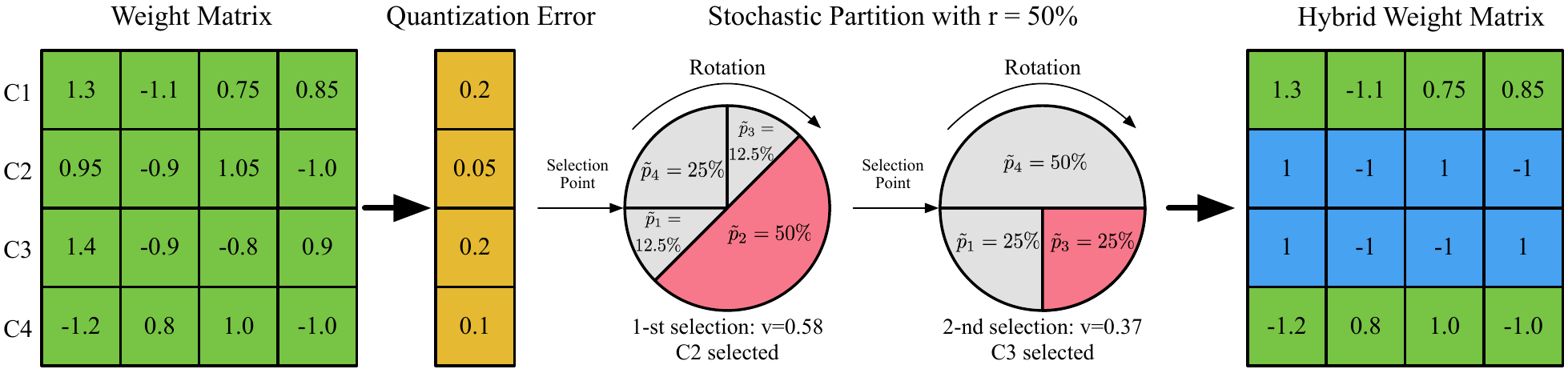}
  \vspace{2ex}
  \caption{Illustration of the stochastic quantization procedure. Given the weight matrix of a conv-layer and a SQ ratio $r$, we first calculate the quantization error.
Then we derive the quantization probability for each filter channel (rows of the weight matrix), and adopt a sampling without replacement roulette algorithm based on the probability to select a portion of quantized rows. Finally, we obtain the weight matrix mixed with quantized and full-precision rows, and perform forward and backward propagation based on this hybrid matrix during training procedure.}
\label{fig:framework}

\end{figure}

In this section, we elaborate the SQ algorithm.
As the low-bit quantization is done layer-by-layer in existing low-bit DNNs \cite{courbariaux2015binaryconnect,rastegari2016xnor,li2016ternary},
we follow the same paradigm. For simplicity, here we just introduce the algorithm for one layer.
We propose to quantize a portion of the weights to low-bits in each layer during forward and backward propagation to minimize the information reduction due to full quantization.
\autoref{fig:framework} illustrates the stochastic quantization algorithm.
We first calculate the quantization error, and then derive a quantization probability $p$ for each element/filter (See Section~\ref{sec:prob} for details).
Given a SQ ratio $r$, we then stochastically select a portion of elements/filters to quantize by a roulette algorithm introduced in Section~\ref{sec:sto}.
We finally demonstrate the training procedure in Section~\ref{sec:train}.

\subsection{Stochastic Partition} \label{sec:sto}
In each iteration, given the weight matrix $\mathcal{W} = \{\mathbf{W}_1, ... ,\mathbf{W}_m\}$ of each layer, we want to partition the rows of $\mathcal{W}$ into two disjoint groups $G_q = \{\mathbf{W}_{q_1}, \cdots, \mathbf{W}_{q_{N_q}}\}$ and $G_r = \{\mathbf{W}_{r_1}, \cdots, \mathbf{W}_{r_{N_r}}\}$, which should satisfy
\begin{equation}
\footnotesize
    G_q \cup G_r = \mathcal{W} \quad \text{and} \quad G_q \cap G_r = \varnothing,
\end{equation}
where $\mathbf{W}_{i} \in \mathbb{R}^d$ is the filter for the \textit{i}-th output channel or $i$-th row of weight matrix $\mathcal{W}$.
We quantize the rows of $\mathcal{W}$ in group $G_q$ to low bitwidth values while keeping the others in group $G_r$ full-precision.
$G_q$ contains $N_q$ items, which is restricted by the SQ ratio $r$ ($N_q = r \times m$). Meanwhile, $N_r = (1-r) \times m$.
The SQ ratio $r$ increases gradually to $100\%$ to make the whole weight matrix $\mathcal{W}$ and the whole network quantized in the end of training.
We introduce a quantization probability $\mathbf{p} \in \mathbb{R}^m$ over each row of $\mathcal{W}$, where the \textit{i}-th element $p_i$ indicates the probability of the \textit{i}-th row to be quantized. The quantization probability is defined over the quantization error, which will be described in Section~\ref{sec:prob}.\footnote{We can also define the non-quantization probability indicating the probability each row should not be quantized.
%Our implementation is based on the non-quantization probability since it requires less selection operations in the \textit{roulette} algorithm due to the fact that $N_r$ is always smaller than $N_q$.
}

Given the SQ ratio $r$ and the quantization probability $\mathbf{p}$ for each layer, we propose a \textit{sampling without replacement roulette} algorithm to select $N_q$ rows to be quantized for group $G_q$, while the remaining $N_r$ rows are still kept full-precision.
Algorithm~\ref{alg:roulette} illustrates the roulette selecting procedure.
%Each channel corresponds to a fan region in the roulette, while the area of fan is determined by the probability $p_j$.  We rotate the roulette by randomly sampling a value from the uniform distribution and select a lucky channel with probability just large than the random value, and put the channel in $G_q$. After each selection, we set $p_j = 0$ to make sure each channel only being selected once. We repeat this selection procedure for $N_q$ times.
%
%The computing complexity of the probability normalization and selection is \textit{O(m)}, so the entire computing complexity of the roulette algorithm is $O(N_q \times m)$. We will introduce some tricks to speed up this algorithm in Section~\ref{sec:details}.
\begin{algorithm}[t]
\footnotesize
\caption{Roulette algorithm to partition weight matrix into quantized and real-valued groups}
\label{alg:roulette}
\begin{algorithmic}[1]
\Require
The SQ ratio $r$ and the quantization probability vector $\mathbf{p} \in \mathbb{R}^m$ over output channels of weight matrix $\mathcal{W}$.
\Ensure
group $G_q$ and $G_r$.
\State $G_q = \varnothing$; $G_r = \varnothing$;
\State $N_q = r \times m$;
\For {$i = 1$ to $N_q$}
\State Normalize $\mathbf{p}$ with $\tilde{\mathbf{p}} = \mathbf{p}/\|\mathbf{p}\|_1$; \Comment{$\|\mathbf{p}\|_1$ is $L_1$ norm of $\mathbf{p}$}
\State Sample a random value $v_i$ uniformly in $(0,1]$;
\State Set $s_i = 0$, and $j = 0$;  \Comment{$s_i$ accumulates the normalized probability}
\While {$s_i < v_i$};
\State $j = j + 1$; $s_i = s_i + \tilde{p}_j$;  \Comment{$\tilde{p}_j$ is the $j$-th element in $\tilde{\mathbf{p}}$}
\EndWhile
\State $G_q = G_q \cup \{\mathbf{W}_j\}$;
\State Set $p_j = 0$;   \Comment{avoid $j$-th channels being selected again}
\EndFor
\State $G_r = \mathcal{W} \setminus G_q$;
\end{algorithmic}
\end{algorithm}

\subsection{From Quantization Error to Quantization Probability}\label{sec:prob}
%The quantization probability measures the extent that a channel should be quantized. The larger the value is, the bigger the probability that the channel should be quantized.
Recall that the motivation of this work is to alleviate the accuracy drop due to the inappropriate gradient direction from large quantization errors,
the quantization probability over channels should base on the quantization error between the quantized and real-valued weights.
If the quantization error is small, quantizing the corresponding channel brings little information reduction, then we should assign a high quantization probability to this channel. That means, the quantization probability should be inversely proportional to the quantization errors.

We generally denote $\mathbf{Q}_i$ as the quantized version of the full-precision weight vector $\mathbf{W}_i$.
For simplicity, we omit the bitwidth or scaling factor $\alpha$ in $\mathbf{Q}_i$.  That means  $\mathbf{Q}_i$ can be $\mathbf{B}_i$, $\mathbf{T}_i$ or even $\alpha\mathbf{B}_i$, $\alpha\mathbf{T}_i$ for different low-bit DNNs.
We measure the quantization error in terms of the normalized $L_1$ distance between $\mathbf{W}_i$ and $\mathbf{Q}_i$:
\begin{equation}
\footnotesize
\label{equ:error}
e_i = \frac{\|\mathbf{W}_i - \mathbf{Q}_i\|_{1}}{\|\mathbf{W}_i\|_{1}}.
\end{equation}
Then we can define the quantization probability given the quantization error $\mathbf{e} = [e_1, e_2, ..., e_n]$.
The quantization probability $p_i$ should be inversely proportional to $e_i$. We define an intermediate variable $f_i = \nicefrac{1}{e_i + \epsilon}$ to represent the inverse relationship, where $\epsilon$ is a small value such as $10^{-7}$ to avoid possible overflow.
The probability function should be a monotonically non-decreasing function of $f_i$. Here we consider four different choices:
\begin{itemize}
\addtolength{\itemsep}{-0.08in}
\item[(1)] A constant function to make equal probability over all channels: $p_i = \nicefrac{1}{m}$. This function ignores the quantization error with a totally random selection strategy.
\item[(2)] A linear function defined as $p_i = \nicefrac{f_i}{\sum_j f_j}$.
\item[(3)] A softmax function defined as $p_i = \nicefrac{\exp (f_i)}{\sum_j \exp (f_j)}$.
\item[(4)] A sigmoid function defined as $p_i = \nicefrac{1}{1 + \exp (-f_i)}$.
\end{itemize}
We will empirically compare the performance of these choices in Section~\ref{sec:ablation}.

\subsection{Training with Stochastic Quantization}\label{sec:train}
When applied the proposed stochastic quantization algorithm to train a low-bit DNN, it will involve four steps: stochastic partition of weights, forward propagation, backward propagation and parameter update. %We only quantize the weights in $G_q$ during forward and backward propagation.

Algorithm~\ref{alg:training} illustrates the procedure for training a low-bit DNN with stochastic quantization.
First, all the rows of $\mathcal{W}$ are quantized to obtain $\mathcal{Q}$.
Note that we do not specify the quantization algorithm and the bitwidth here, since our algorithm is flexibly applied to all the cases.
We then calculate the quantization error by Eq.~\eqref{equ:error} and the corresponding quantization probability.
Given these parameters, we use Algorithm~\ref{alg:roulette} to partition $\mathcal{W}$ into $G_q$ and $G_r$.
We then form the hybrid weight matrix $\mathcal{\widetilde{Q}}$ composed of real-valued weights and quantized weights based on the partition result. If $\mathbf{W}_i \in G_q$, we use its quantized version $\mathbf{Q}_i$ in $\mathcal{\widetilde{Q}}$. Otherwise, we use $\mathbf{W}_i$ directly.
$\mathcal{\widetilde{Q}}$ approximates the real-valued weight matrix $\mathcal{W}$ much better than $\mathcal{Q}$, and thus provides much more appropriate gradient direction.
We update $\mathcal{W}$ with the hybrid gradients $\frac{\partial \mathcal{L}}{\partial \mathcal{\widetilde{Q}}^t}$ in each iteration as
\begin{equation}
\footnotesize
\mathcal{W}^{t+1} = \mathcal{W}^t - \eta^t \frac{\partial \mathcal{L}}{\partial \mathcal{\widetilde{Q}}^t},
\label{eq:update}
\end{equation}
where $\eta^t$ is the learning rate at $t$-th iteration. 
That means the quantized part is updated with gradients derived from quantized weights, while the real-valued part is still updated with gradients from real-valued weights.
Lastly, the learning rate and SQ ratio get updated with a pre-defined rule. We will specify this in Section~\ref{sec:details}.

When the training stops, $r$ increases to $100\%$ and all the weights in the network are quantized. There is no need to keep the real-valued weights $\mathcal{W}$ further.
We perform forward propagation with low bitwidth weights $\mathcal{Q}$ during the testing phase.
\begin{algorithm}[t]
\footnotesize
\caption{Training algorithm based on SQ}
\label{alg:training}
\begin{algorithmic}[1]
\Require
    \Statex Mini-batch of inputs and targets \{$\mathbf{X}$, $\mathbf{Y}$\}, loss function $\mathcal{L}(\mathbf{Y}, \hat{\mathbf{Y}})$;
    
    \Statex weights $\mathcal{W}^t$, learning rate $\eta^t$ and SQ ratio $r^t$ of $t$-th iteration.
\Ensure
Updated parameters $\mathcal{W}^{t+1}$, learning rate $\eta^{t+1}$ and SQ ratio $r^{t+1}$.
\State Quantize each row of $\mathcal{W}^t$, and obtain quantized matrix $\mathcal{Q}^t$;
\If{$r^t < 100\%$}
\State Calculate the quantization error $\mathbf{e}$;
\State Calculate the quantization probability $\mathbf{p}$;
\State Partition $\mathcal{W}^t$ into $G_q$ and $G_r$ by Algorithm~\ref{alg:roulette} with $r^t$ and $\mathbf{p}$;
\State Form the hybrid weight matrix $\mathcal{\widetilde{Q}}^t$, where each row $\mathbf{\widetilde{Q}}_i = \mathbf{W}_i$ \textit{if} $\mathbf{W}_i \in G_r$; \textit{else} $\mathbf{\widetilde{Q}}_i = \mathbf{Q}_i$;
\Else
\State $\mathcal{\widetilde{Q}}^t = \mathcal{Q}^t$;
\EndIf
\State $\hat{\mathbf{Y}}$ = \textbf{Forward}($\mathbf{X}$, $\mathcal{\widetilde{Q}}^t$); \Comment{Forward to get the target estimation}
\State $\frac{\partial \mathcal{L}}{\partial \mathcal{\widetilde{Q}}^t}$ = \textbf{Backward}($\frac{\partial \mathcal{L}}{\partial \hat{\mathbf{Y}}}$, $\mathcal{\widetilde{Q}}^t$); \Comment{Backward to get the gradient of $\mathcal{\widetilde{Q}}^t$}
\State Update $\mathcal{W}^{t+1}$ according to Eq.~\eqref{eq:update}.
\State $\eta^{t+1}, r^{t+1}$ = \textbf{Update}($\eta^t, r^t, t$);
\end{algorithmic}
\end{algorithm}

\section{Experiments}
We conduct extensive experiments on the CIFAR-10, CIFAR-100 and ImageNet large scale classification datasets to validate the effectiveness of the proposed algorithm. We apply the SQ algorithm on two kinds of low-bit settings (\textit{aka} BWN and TWN), and show the advantages over the existing low-bit DNN algorithms.

\subsection{Datasets and Implementation Details}\label{sec:details}
We briefly introduce the datasets used in our experiments, including the corresponding network structures and training settings. 

\textbf{CIFAR}~\cite{krizhevsky2009learning} consists of a training set of $50,000$ and a test set of $10,000$ color images of resolution $32 \times 32$ with $10$ classes in CIFAR-10 and $100$ classes in CIFAR-100.
We adopt two network architectures. The first one is derived from VGGNet~\cite{simonyan2014very} and denoted as VGG-9~\cite{courbariaux2015binaryconnect}, by the architecture of ``$(2\times 64C3) - MP2 - (2\times 128C3) - MP2 - (2\times 256C3) - MP2 - (2\times 512FC) - 10/100FC - Softmax$''. 
The network is trained with a SGD solver with momentum $0.9$, weight decay $0.0001$ and batch size $100$. 
We do not use any data augmentation in this network. 
We start with a learning rate $0.1$, divide it by $10$ after each $15$k iterations, and terminate the training at $100$k iterations. 
The second network is the ResNet-56 architecture with the same training settings as~\cite{he2015deep}.

\textbf{ImageNet}~\cite{russakovsky2015imagenet} is a large scale classification dataset. We use the ILSVRC 2012 dataset in the experiments which consists $1.28$ million training images of $1000$ classes. We evaluate the performance on the $50$K validation images. We adopt a variant of AlexNet~\cite{krizhevsky2012imagenet} architecture with batch normalization (AlexNet-BN) and the ResNet-18~\cite{he2015deep} architecture.
We train the AlexNet-BN by SGD with momentum $0.9$, weight decay $5\times 10^{-5}$ and batch size $256$. The initial learning rate is $0.01$, which is divided by $10$ each after $100$k, $150$k and $180$k iterations. The training is terminated after $200$k iterations. We adopt batch normalization in AlexNet for faster convergence and better stability.
ResNet-18 is trained with a SGD solver with batch size $100$ and the same momentum and weight decay values as AlexNet-BN. The learning rate starts from $0.05$ and is divided by $10$ when the error plateaus. We train it for $300$k iterations.

We implement our codes based on the Caffe framework~\cite{jia2014caffe}, and make training and testing on NVidia Titan X GPU. 
We make full experimental comparison between full-precision weight networks (FWN), Binary Weighted Networks (BWN), and Ternary Weighted Networks (TWN).
We denote BWN and TWN trained with the proposed SQ algorithm as SQ-BWN and SQ-TWN respectively.
For CIFAR-10 and CIFAR-100, we compare the results of FWN, BWN, TWN, SQ-BWN and SQ-TWN on two network architectures VGG-9 and ResNet-56.
For ImageNet, we also make experimental comparison between these five settings with the AlexNet-BN and ResNet-18 network architectures. 

The \textit{SQ ratio $r$} is one critical hyper-parameter in our algorithm, which should be gradually increased to $100\%$ to make all the network weights quantized.
In this work, we divide the training into several stages with a fixed $r$ at each stage.
When the training is converged in one stage, we increase $r$ and move to the next stage, until $r = 100\%$ in the final stage.
%Although we could use the adaptive methods to set $r$ for each iteration, we leave it to future work.
%We find the proposed simple strategy works well in practice.
We conduct ablation study on the scheme for updating the SQ ratio in Section~\ref{sec:ablation}.
We make another simplification that $r$ is the same for all layers in a DNN within one training stage.
When applied the SQ algorithm to train low-bit DNNs, the number of training iterations is $s\times$ of original low-bit DNNs, where $s$ is the number of SQ stages and we use the same number of iterations as other low-bit DNNs (\textit{i.e.}, BWN, TWN) in each stage (although we do not need so many iterations to converge).

\subsection{Ablation Study}\label{sec:ablation}
There are several factors in the proposed method which will affect the final results, including selection granularity, partition algorithm, quantization probability function and scheme for updating the SQ ratio.
Therefore, we design a series of experiments for each factor and analyze their impacts to our algorithm.
All the ablation studies are based on the ResNet-56 network and the CIFAR-10 dataset.
In consideration of the interaction between these factors, we adopt the variable-controlling approach to ease our study. The default settings for selection granularity, partition algorithm, quantization probability function and  SQ ratio scheme are channel-wise selection, stochastic partition (roulette), linear function and four stages training with the SQ ratio $r = 50\%, 75\%, 87.5\%$ and $100\%$.

\begin{table}[]
\footnotesize
\begin{minipage}{.42\textwidth}
\centering
\begin{tabular}[0.90\columnwidth]{c|c|c}
\hline
& Channel-wise & Element-wise\\
\hline\hline
SQ-BWN & \textbf{7.15} & 7.67 \\
SQ-TWN & \textbf{6.20} & 6.53 \\
\hline
\end{tabular}
\vspace{2ex}
\caption{Test error ($\%$) of selection granularity on CIFAR-10 with ResNet-56.}
\label{tab:granularity}
\end{minipage}
\hfill
\begin{minipage}{.53\textwidth}
\centering
\begin{tabular}[0.90\columnwidth]{c|c|c|c}
\hline
& Stochastic & Deterministic & Fixed\\
\hline\hline
SQ-BWN & \textbf{7.15} & 8.21 & *\\
SQ-TWN & \textbf{6.20} & 6.85 & 6.50\\
\hline
\end{tabular}
\vspace{2ex}
\caption{Test error ($\%$) of different partition algorithms on CIFAR-10 with ResNet-56. * means not converged.}
\label{tab:selection}
\end{minipage}
\end{table}

\begin{table}[]
\footnotesize
\begin{minipage}{.45\textwidth}
\centering
\begin{tabular}[0.8\columnwidth]{c|c|c|c}
\hline
& Exp & Ave & Tune\\
\hline\hline
SQ-BWN & \textbf{7.15} & 7.35 & 7.18\\
SQ-TWN & \textbf{6.20} & 6.88 &  6.62\\
\hline
\end{tabular}
\vspace{2.0ex}
\caption{Test error rates ($\%$) of different schemes for updating SQ ratio on CIFAR-10 with ResNet-56.}
\label{tab:stage}
\end{minipage}
\hfill
\begin{minipage}{.52\textwidth}
\centering
\begin{tabular}[0.8\columnwidth]{c|c|c|c|c}
\hline
& Linear & Constant & Softmax & Sigmoid\\
\hline\hline
SQ-BWN & \textbf{7.15} & 7.44 & 7.51 & 7.37\\
SQ-TWN & \textbf{6.20} & 6.30 & 6.29 & 6.28\\
\hline
\end{tabular}
\vspace{2.0ex}
\caption{Test error ($\%$) of different probability functions given quantization error on  CIFAR-10 with ResNet-56.}
\label{tab:function}
\end{minipage}
\vspace{-1.0ex}
\end{table}

\vspace{-2ex}
\paragraph{Channel-wise \textit{v.s.} Element-wise.}
We first study the impact of quantization granularity on performance. We adopt two strategies, \textit{i.e.}, channel-wise and element-wise, with the results shown in~\autoref{tab:granularity}. Element-wise selection ignores the filter structure and the interactions within each channel, which leads to lower accuracy. By treating each filter-channel as a whole, we can preserve the structures in channels and obtain higher accuracy.

\vspace{-2ex}
\paragraph{Stochastic \textit{v.s.} Deterministic \textit{v.s.} Fixed.}
The proposed algorithm uses stochastic partition based on the roulette algorithm to select $N_q$ rows to be quantized.
We argue that the stochastic partition can eliminate the requirement of finding the best partition after initialization, and have the ability to explore the searching space for a better solution due to the exploitation-exploration nature.
To prove that, we compare our pure stochastic scheme to another two none fully stochastic schemes.
The first one is the deterministic scheme using sorting based partition.
Given the SQ ratio $r$ and the quantization error $\mathbf{e}$, we sort rows of weight matrix based on $\mathbf{e}$ and select $N_q$ rows with the least quantization error as group $G_q$ in each iteration.
The second one is the fixed partition, in which we only do roulette partition for the first iteration, and keep $G_q$ and $G_r$ fixed for all the iterations in each training stage.

\autoref{tab:selection} compares the results. It shows that the stochastic algorithm gets significantly better results in BWN and TWN, which proves that the stochastic partition we introduced is a key factor for success. Another interpolation is that stochastic quantization acts as an regularizer in training, which also benefits the performance.

\vspace{-2ex}
\paragraph{Quantization Probability Function.}
In Section~\ref{sec:prob}, we introduce four choices of quantization probability function given the quantization error, which are constant, linear, softmax and sigmoid functions. We compare the results of different functions in~\autoref{tab:function}. Among these choices, linear function beats all other competitors due to its better performance and simplicity. We find that the performance of different functions are very close, which indicates that what matters most is the stochastic partition algorithm itself.

\vspace{-2ex}
\paragraph{Scheme for updating the SQ ratio.}
We also study how to design the scheme for updating the SQ ratio. Basically, we assume that training with stochastic quantization will be divided into several stages, and each stage has a fixed SQ ratio.
In all previous setting, we use four stages with SQ ratio $r = 50\%, 75\%, 87.5\%$ and $100\%$. We call this exponential scheme (Exp) since
in each stage, the non-quantized part is half-sized to that of previous stage.
%The intuition behind the exponential update is that the remaining real-valued weights become more important when their size reduces, so we need to move smaller steps gradually.
We compare our results with two additional schemes: the average scheme (Ave), and the fine-tuned scheme (Tune).
The average scheme includes five stages with $r = 20\%, 40\%, 60\%, 80\%$ and $100\%$.
The fine-tuned scheme tries to fine-tune the pre-trained full-precision models with stages same as the exponential scheme instead of training from scratch.

\autoref{tab:stage} shows the results of the compared three schemes.
It is obvious that the exponential scheme performs better than the average scheme and the fine-tuned scheme.
It can also be concluded that our method works well when training from scratch.

\begin{table}[]
\centering
\footnotesize
%\resizebox{0.6\textwidth}{!}{
\begin{tabular}{c|c|c|c|c|c}
\hline
\multirow{2}{*}{} & \multirow{2}{*}{Bits} & \multicolumn{2}{c|}{CIFAR-10} & \multicolumn{2}{c}{CIFAR-100}\\
\cline{3-6}
& & VGG-9 & ResNet-56 & VGG-9 & ResNet-56\\
\hline\hline
FWN & 32 & 9.00 & 6.69 & 30.68 & 29.49 \\
\hline
BWN & 1 & 10.67 & 16.42 & 37.68 & 35.01 \\
SQ-BWN & 1 & 9.40 & 7.15 & 35.25 & 31.56 \\
\hline
TWN & 2 & 9.87 & 7.64 & 34.80 & 32.09 \\
SQ-TWN & 2 & \textbf{8.37} & \textbf{6.20} & 34.24 & \textbf{28.90} \\
\hline
\end{tabular}
%}
\vspace{2.0ex}
\caption{Test error ($\%$) of VGG-9 and ResNet-56 trained with 5 different methods on the CIFAR-10 and CIFAR-100 datasets. Our SQ algorithm can consistently improve the results. SQ-TWN even outperform full-precision models.}
\label{tab:cifar}
\end{table}

\begin{table}[]
\centering
\footnotesize
%\resizebox{0.6\textwidth}{!}{
\begin{tabular}{c|c|c|c|c|c}
\hline
\multirow{2}{*}{} & \multirow{2}{*}{Bits} & \multicolumn{2}{c|}{AlexNet-BN} & \multicolumn{2}{c}{ResNet-18}\\
\cline{3-6}
& & top-1 & top-5 & top-1 & top-5\\
\hline\hline
FWN & 32 & 44.18 & 20.83 & 34.80 & 13.60 \\
\hline
BWN & 1 & 51.22 & 27.18 & 45.20 & 21.08 \\
SQ-BWN & 1 & 48.78 & 24.86 & 41.64 & 18.35 \\
\hline
TWN & 2 & 47.54 & 23.81 & 39.83 & 17.02 \\
SQ-TWN & 2 & 44.70 & 21.40 & 36.18 & 14.26 \\
\hline
\end{tabular}
%}
\vspace{2.0ex}
\caption{Test error ($\%$) of AlexNet-BN and ResNet-18 trained with 5 different methods on the ImageNet dataset.}
\label{tab:imagenet}
\vspace{-1ex}
\end{table}

\subsection{Benchmark Results}\label{sec:results}
We make a full benchmark comparison on the CIFAR-10, CIFAR-100 and ImageNet datasets between the proposed SQ algorithm and traditional algorithms on BWN and TWN.
For fair comparison, we set the learning rate, optimization method, batch size etc identical in all these experiments.
We adopt the best setting (\textit{i.e.}, channel-wise selection, stochastic partition, linear probability function, exponential scheme for SQ ratio) for all SQ based experiments.

\autoref{tab:cifar} presents the results on the CIFAR-10 and CIFAR-100 datasets. In these two cases, SQ-BWN
and SQ-TWN outperform BWN and TWN significantly, especially in ResNet-56 which are
deeper and the gradients can be easily misled by quantized weights with large quantization
errors. For example, SQ-BWN improve the accuracy by $9.27\%$ than BWN on the CIFAR-10 dataset
with the ResNet-56 network. The 2-bits SQ-TWN models can even obtain higher accuracy than the
full-precision models. Our results show that SQ-TWN improve the accuracy by $0.63\%$,
$0.49\%$ and $0.59\%$ than full-precision models on CIFAR-10 with VGG-9, CIFAR-10 with ResNet-56
and CIFAR-100 with ResNet-56, respectively.

\begin{minipage}[t]{\textwidth}
\hspace{-2ex}
\begin{minipage}[b]{.47\textwidth}
\centering
\includegraphics[width=0.99\columnwidth]{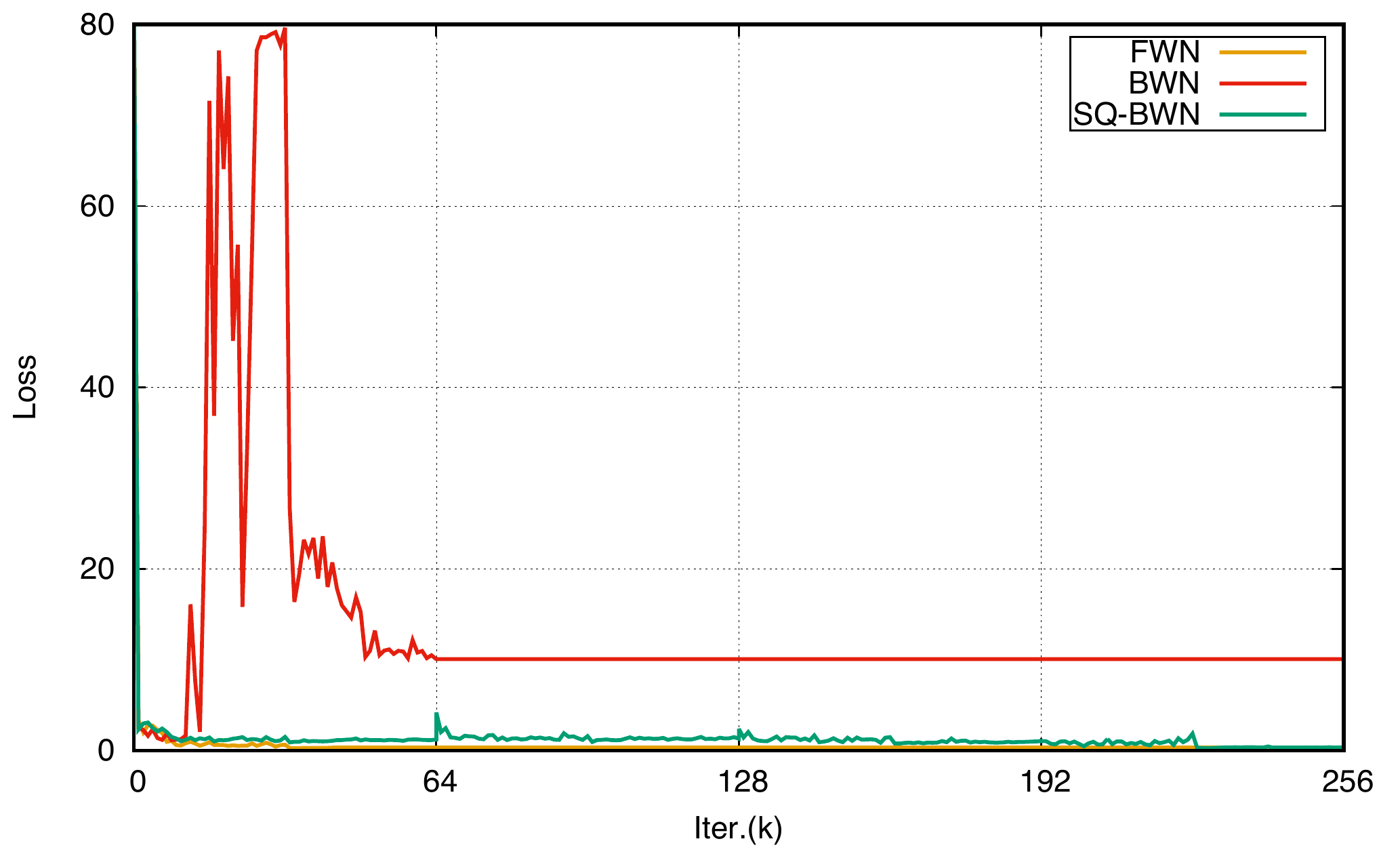}
\vspace{0.5ex}
\captionof{figure}{Test loss of FWN, BWN and SQ-BWN on CIFAR-10 with ResNet-56.}
\label{fig:loss-BWN}
\end{minipage}
\hspace{1ex}
\begin{minipage}[b]{.47\textwidth}
\centering
\includegraphics[width=0.99\columnwidth]{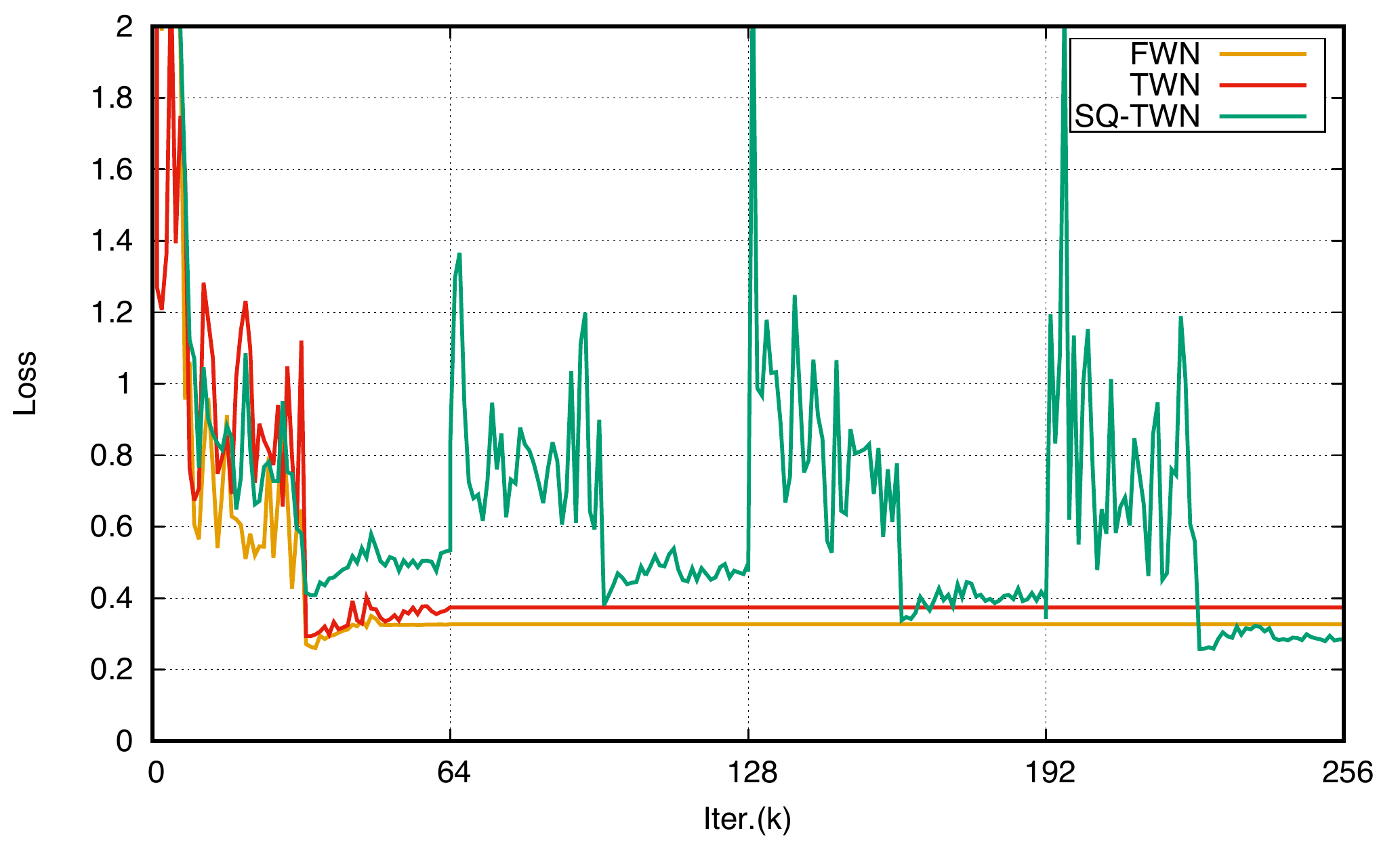}
\vspace{0.5ex}
\captionof{figure}{Test loss of FWN, TWN and SQ-TWN on CIFAR-10 with ResNet-56.}
\label{fig:loss-TWN}
\end{minipage}
\vspace{4ex}
\end{minipage}

In~\autoref{fig:loss-BWN} and~\autoref{fig:loss-TWN}, we show the curves of test loss on the CIFAR-10 dataset with the ResNet-56 network. In~\autoref{fig:loss-BWN}, BWN doesn't converge well with much larger loss than SQ-BWN, while the losses in SQ-BWN and FWN are relatively low. In~\autoref{fig:loss-TWN}, we can see that at the beginning of each stage, quantizing more weights leads to large loss, which will be soon converged after some training iterations. Finally, SQ-TWN gets smaller loss than FWN, while the loss in TWN is larger than FWN.

\autoref{tab:imagenet} shows the results on the standard ImageNet 50K validation set. We compare SQ-BWN, SQ-TWN to FWN, BWN and TWN with the AlexNet-BN and the ResNet-18 architectures, and all the errors are reported with only single center crop testing.
It shows that our algorithm helps to improve the performance quite a lot, which consistently beats the baseline methods by a large margin. We can also see that SQ-TWN yield approaching results with FWN.
Note that some works \cite{rastegari2016xnor,venkatesh2016accelerating} keep the first or last layer full-precision to alleviate possible accuracy loss. Although we did not do like that, SQ-TWN still achieves near full-precision accuracy. Nevertheless, we conclude that our SQ algorithm can consistently and significantly improve the performance.

\section{Conclusion}
In this paper, we propose a Stochastic Quantization (SQ) algorithm to learn accurate low-bit DNNs.
We propose a roulette based partition algorithm to select a portion of weights to quantize, while keeping the other portion unchanged with full-precision, at each iteration. The hybrid weights provide much more appropriate gradients and lead to better local minimum. We empirically prove the effectiveness of our algorithm by extensive experiments on various low bitwidth settings, network architectures and benchmark datasets.
We also make our codes public at \url{https://github.com/dongyp13/Stochastic-Quantization}.
Future direction may consider the stochastic quantization of both weights and activations.

\section*{Acknowledgement}
This work was done when Yinpeng Dong and Renkun Ni were interns at Intel Labs supervised by Jianguo Li. Yinpeng Dong, Jun Zhu and Hang Su are also supported by the National Basic Research Program of China (2013CB329403), the National Natural Science Foundation of China (61620106010, 61621136008).

\bibliography{egbib}
\end{document}